\documentclass[10pt,twocolumn,twoside]{IEEEtran}

\usepackage{graphicx}
\usepackage{cite}
\usepackage{amssymb}
\usepackage{amsfonts}
\usepackage{url}
\usepackage{color,xcolor}
\begin{document}
%
\title{Line Drawings of Natural Scenes Guide Visual Attention}

\author{Kai-Fu~Yang, Wen-Wen~Jiang, Teng-Fei~Zhan, and~Yong-Jie~Li,~\IEEEmembership{Member,~IEEE}
\thanks{Kai-Fu Yang, Wen-Wen Jiang, Teng-Fei Zhan, and Yong-Jie Li are with MOE Key Lab for Neuroinformation，University of Electronic Science and Technology of China, Chengdu, China.}}

\markboth{}%
{}

\IEEEcompsoctitleabstractindextext{%
\begin{abstract}
Visual search is an important strategy of the human visual system for fast scene perception. The guided search theory suggests that the global layout or other top-down sources of scenes play a crucial role in guiding object searching. In order to verify the specific roles of scene layout and regional cues in guiding visual attention, we executed a psychophysical experiment to record the human fixations on line drawings of natural scenes with an eye-tracking system in this work. We collected the human fixations of ten subjects from 498 natural images and of another ten subjects from the corresponding 996 human-marked line drawings of boundaries (two boundary maps per image) under free-viewing condition. The experimental results show that with the absence of some basic features like color and luminance, the distribution of the fixations on the line drawings has a high correlation with that on the natural images. Moreover, compared to the basic cues of regions, subjects pay more attention to the closed regions of line drawings which are usually related to the dominant objects of the scenes. Finally, we built a computational model to demonstrate that the fixation information on the line drawings can be used to significantly improve the performances of classical bottom-up models for fixation prediction in natural scenes. These results support that Gestalt features and scene layout are important cues for guiding fast visual object searching.
\end{abstract}

}

\maketitle
\IEEEdisplaynotcompsoctitleabstractindextext
\IEEEpeerreviewmaketitle

\section{Introduction}
\IEEEPARstart{A}{ttention} via visual search is necessary for rapid scene perception and object searching because information processing in the visual system is limited to one or a few targets or regions at one time \cite{wolfe2011visual}. From the viewpoint of engineering, modelling visual search usually facilitates the subsequent higher visual processing tasks, such as image re-targeting \cite{goferman2012context}, image compression \cite{christopoulos2000jpeg2000}, object recognition \cite{rutishauser2004bottom}, etc. However, how visual attention is guided in real visual searching tasks is an interesting but un-fully solved question in the field of visual neuroscience.

Studies of physiology and psychophysics have revealed that several sources of factors play important roles in guiding visual attention. The first one is bottom-up guidance. The classical Feature Integration Theory (FIT) suggests that visual attention is driven by some low-level visual cues such as local luminance, color and orientation \cite{treisman1980feature, koch1987shifts}. With a bottom-up manner, these cues are processed in parallel channels and combined to guide the visual attention. This stimulus-driven guidance causes that some local regions with feature difference in scenes attract more attention than others. Moreover, the studies of computational models based on the FIT (e.g., IT \cite{itti1998model, itti2001computational}) also show that these models employing local cues usually predict the attention limited in some regions with high local contrasts (e.g., object boundaries). 

On the other hand, visual attention is also guided by task related prior knowledge beyond the low-level visual cues \cite{xu2014predicting}. Imagining that we are searching an object in a complex scene, our attention is guided by current task and directed to the objects with known features of the desired targets. Previous studies have revealed that the feedback pathways from high-level cortexes carry rich and varying information about the behavioural context, including visual attention, memory, expectation, etc. \cite{gilbert2013top}. For example, the classical study by Yarbus \cite{yarbus1967eye} shows that human fixation distribution is dependent on the asked question when viewing a picture. In addition, user- or task-driven guidance is also widely used in visual processing models in computer vision area \cite{navalpakkam2005modeling, Kanan2009SUN, borji2014look}. Recent interesting works \cite{eckstein2017humans, wolfe2017visual} show that human attention is guided by the knowledge of the plausible size of target objects. It was suggested that this is a useful brain strategy for human to rapidly discount the distracters (e.g., the object with atypical size relative to the surrounding) during visual searching, and this is neglected in current machine learning systems \cite{eckstein2017humans}. 

Besides the stimulus-driven and task-driven guidance, contextual guidance is another important source for guiding general visual searching tasks \cite{torralba2006contextual}. According to the Guided Search Theory (GST) \cite{wolfe1994guided}, the scene context and global statistical information are important for guiding visual object searching. This indicates that rapid global scene analysis will facilitate the prediction of the locations of potential objects. The GST suggests that rapid visual searching involves two parallel pathways \cite{wolfe2011visual}: (1) the non-selective pathway, which serves to extract spatial layout (or gist) information rapidly from the entire scenes; and (2) the selective pathway, which works to extract and bind the low-level features under the guidance of the contextual information of scene extracted from the non-selective pathway. This two-pathway based strategy provides a unified framework to integrate the context of scene and the local information for rapid visual searching. Recently, we have built a computational model based on GST to effectively execute the general task of salient structure detection in natural scenes \cite{yang2016unified}, which further supports the efficiency of context guidance in rapid visual searching with a task-independent manner. However, it has been revealed that such context guidance also employs the prior knowledge or experience in our memory and guides the low-level feature integration with a top-down manner, similar to the task-driven guidance \cite{eckstein2011visual, bar2004visual}. 

According to the source of contextual information, we can simply classify the contextual guidance into two categories: object-based and scene-based guidance. Nuthmann et al. found a preferred viewing location close to the center of objects within natural scenes when directed with different task instructions, which suggests that visual attentional selection in scenes is object-based \cite{Nuthmann2010Object}. A related biologically plausible model was proposed to attend to the proto-objects in natural scenes \cite{Walther2006Modeling}. In addition, the object-based guidance also indicates that meaningful visual objects usually possess some principles about the shape. For example, the well-known Gestalt theory \cite{Koffka1935Principles} summarizes some universal principles of visual perception (e.g., closure and symmetry), which are crucial factors for facilitating the visual object searching. Based on the object-based prior, Kootstra et al. employed the feature of symmetry for fixation prediction and object segmentation in complex scenes \cite{kootstra2008paying, kootstra2010using}. Yu et al. used the Gestalt grouping cues to select the salient regions from over-segmented scenes \cite{yu2016computational}. 

As for the scene-based guidance, previous studies have shown that structural scene cues, such as layout, horizontal line, openness, depth, etc., usually guide the visual searching \cite{oliva2001modeling, ross2009estimating}. Beyond the geometry of the scenes, visual attention is also guided by the semantic information of the real-world scenes, such as the gist of the scenes \cite{oliva2001modeling}, scene-object relations, and object-object relations \cite{wu2014guidance,boettcher2018anchoring}. Therefore, clear understanding of the neural basis of scene perception, object perception, perceptual learning and memory will also facilitate the understanding of visual searching \cite{peelen2014attention}.

Moreover, scene structure could provide specific scene-related information for specific visual tasks. Recent studies have addressed the role of context in driving \cite{deng2016does, deng2017learning}. By analysing the fixation data of 40 non-drivers and experienced drivers when viewing 100 traffic images, Deng et al. showed that drivers' attention was mostly concentrated on the vanishing points of the roads \cite{deng2016does, deng2017learning}. In addition, their models further support that vanishing point information is more important than other bottom-up cues in guiding drivers' attention during driving. At the same time, Borji et al. also found that vanishing points can guide the attention in free-viewing and visual searching tasks \cite{borji2016vanishing}. 

On the other hand, physiological evidences have shown that the global contours delineating the outlines of visual objects could be responded quite early by the neurons of high visual cortexes, which can provide the sufficient feedback modulation that enhances the object-related responses at the lower visual levels \cite{Poort2012The,Chen2014Incremental}. Therefore, in this paper, we focus on the role of line drawings in visual guidance, which is closely related to the important contextual guidance cues such as regional shapes and scene layout. In specific, we explore the role of line drawings in visual searching with a psychophysical experiment. We first collected the human fixations from 498 natural images and the corresponding 996 human-marked line drawings (two line drawings for each image) with a free-viewing manner. The experimental results show that with the absence of some basic features like color and luminance, the distribution of fixations on line drawings has high correlation with that on natural images. Moreover, the subjects pay more attention to the closed regions of line drawings that are usually related to the dominant objects in the scenes. We also built a computational model to demonstrate that the information of line drawings can be used to significantly improve the performance of some classical bottom-up models for fixation prediction in natural scenes.

\section{Material and Methods}

\subsection{Stimuli and Observers}

We employed 498 natural images from the Berkeley Segmentation Data Set and Benchmarks 500 (BSDS500) \cite{arbelaez2011contour} and collected the eye fixations of 10 subjects on the natural images with a free-viewing manner. This eye-tracking experiment has been conducted in our previous work \cite{yang2015segmentation}. In this work, with the help of the human-marked segmentations and the contour maps (i.e., the line drawings) provided by BSDS500 \cite{arbelaez2011contour}, we further collected the human fixations of another 10 subjects when freely viewing the line drawings. 

In this work, we chose two contour maps as visual stimuli from the 5-10 human-marked contour maps for each natural scene available in BSDS500: the one containing the most contour pixels (denoted by $C_{max}$) and the one containing the least contour pixels (denoted by $C_{min}$). Such choice helps demonstrate the consistent role of contours in guiding visual searching no matter the detailed parts are contained or not in the contour maps. Moreover, in order to keep consistent with our previous experiment with the natural scenes \cite{yang2015segmentation}, each image was resized to 1024$ \times $768 pixels in this work by adding gray margins around them while maintaining the aspect ratio.
The observers include 5 males and 5 females with the mean age of 23.8 (22-25) years old. The selected observers have normal or corrected-to-normal vision for participation. They were naive to the purpose of the experiment and had not previously seen the stimuli. This study was carried out in accordance with the recommendations of “the Guideline for Human Behaviour Studies, the Institutional Review Board of UESTC MRI Research Center for Brain Research with written informed consent from all subjects. All subjects gave written informed consent in accordance with the Declaration of Helsinki. The protocol was approved by the Institutional Review Board of UESTC MRI Research Center for Brain Research.

\begin{figure*}[htbp]
\begin{center}
\includegraphics[width=14cm]{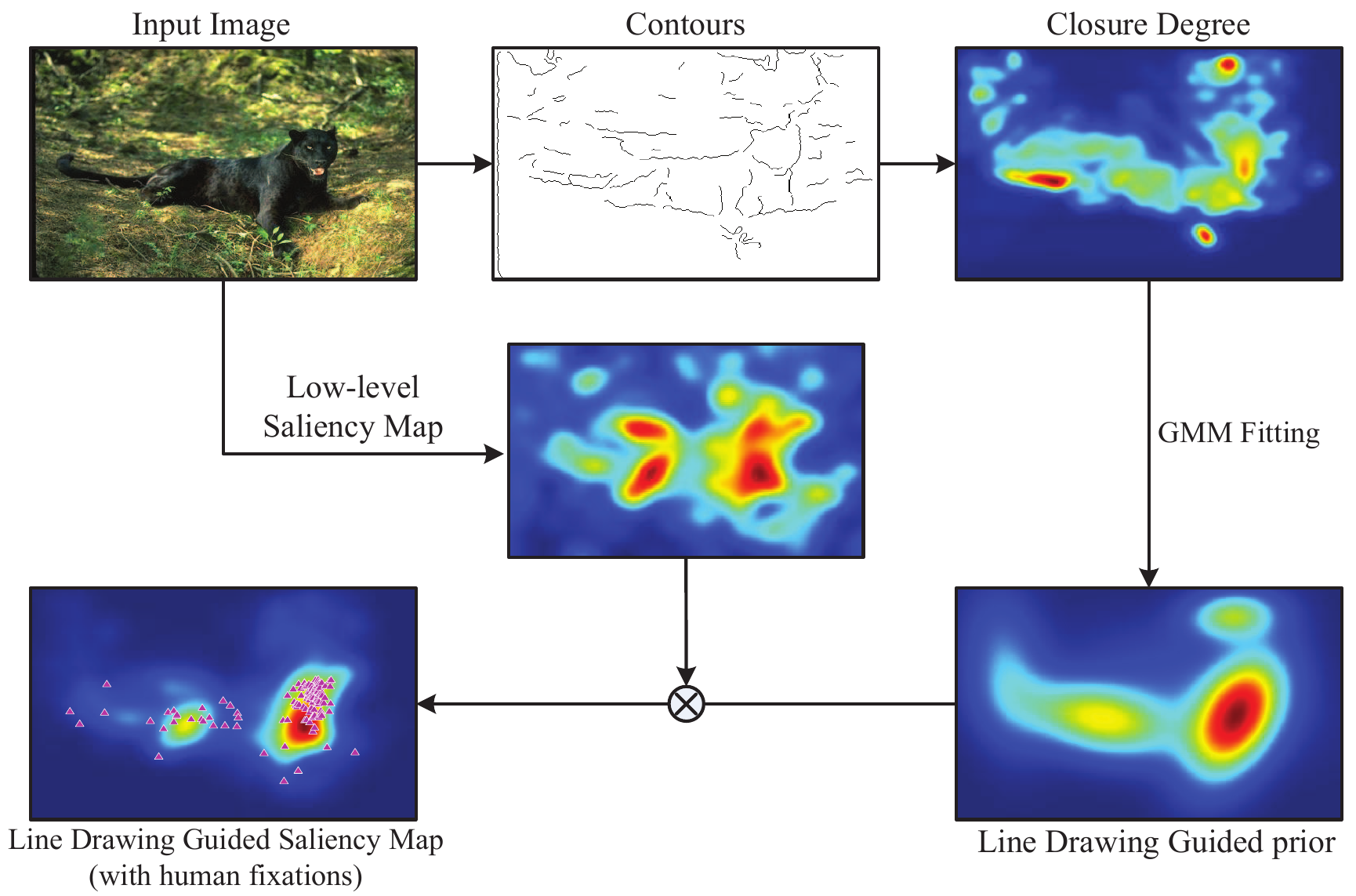}
\end{center}
\caption{The flowchart of the visual attention model guided by line drawings.}
\label{FigModel}
\end{figure*}

\subsection{Procedure}
The procedure is same as that in our previous experiment on the corresponding natural scenes \cite{yang2015segmentation}. Ten subjects were asked to watch the given images displayed on the screen with a free-viewing manner. Each image was presented for 5 seconds after a visual cue of cross (``+") in the center of the gray screen. An Eyelink-2000 eye-tracking system was used to collect the fixations of subjects with a sampling rate of 1000Hz, and the eye-tracking re-calibration was performed on every 30 images. A chin rest was used to stabilize the head.

The procedure of the experiment was designed and programmed based on the Psychotoolbox of MATLAB. The images were presented on the display in a random order, and the contour maps of the same natural images were separated by at least 30 trails. The observers were instructed to simply watch and enjoy the pictures (i.e., free viewing task). Finally, we obtained the fixations of 10 subjects on the line drawings of 498 natural images from the BSDS500 dataset \cite{arbelaez2011contour}. Note that two images of BSDS500 were failed to collect reasonable fixation data because the sizes of the objects in those two images are too small. 



\subsection{Model-based analysis}
In the field of computer vision, numerous saliency detection methods have been proposed, some of which show promising performance, especially when deep learning technology is used to predict human fixations \cite{borji2013state}. However, the researchers in the field of neuroscience devote themselves to make progress to understand how human execute the efficient visual searching. Some researchers have proposed various biologically inspired models for fixation prediction \cite{itti1998model, itti2001computational}. These methods usually have bottom-up architectures and respond to the regions with local high contrast in various feature channels.  

Our experimental results suggest that the closed regions have higher possibility to form meaningful objects and may attract most of our visual fixations (See Results). Therefore, we build a simple model to extract the closed regions and generate the prior location where the objects possibly present. This prior will be used to guide the generation of bottom-up saliency and improve fixation prediction. Figure \ref{FigModel} shows the flowchart of the guided visual attention model.

\begin{figure}[htbp]
\begin{center}
\includegraphics[width=5cm]{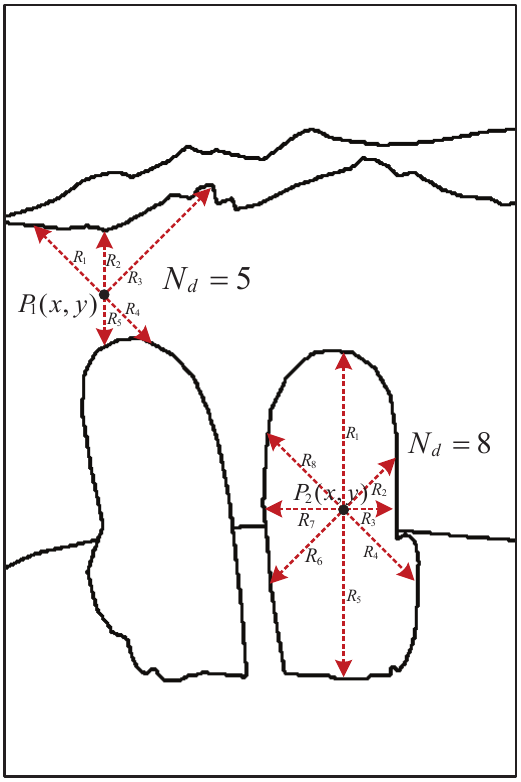}
\end{center}
\caption{The examples of computing closure degree for each point.}
\label{FigClosure}
\end{figure}

Firstly, we extract the edges from the input image with the Structured Forests method presented in \cite{dollar2013structured} and group the main edges into contour with the edge link method \cite{KovesiMATLABCode}. Then, we estimate the possibility of each point in the image within a closed region based on the contour map. Figure \ref{FigClosure} shows two examples of closure computation at different points. For each point $ (x,y) $, we search the contour pixels around it along $ D $ directions (we sample D directions in this study, and $ D=8 $) starting from $ (x,y) $. We use $ N_d(x,y) $ to denote the number contour pixels around (x,y) in all the searched directions. For example, $ P_1(x,y) $ is with $ N_d=5 $, while $ P_2(x,y) $ is with $ N_d=8 $. That means point $ P_2(x,y) $ has the higher possibility of locating in a closed region than $ P_1(x,y) $. 

Computationally, the possibility of $ (x,y) $ within a closed region (denoted by Closure Degree) is obtained with
\begin{equation}
C(x,y) = exp(N_d(x,y)-D)
\label{eqn1}
\end{equation}

\begin{figure*}[htbp]
\begin{center}
\includegraphics[width=12cm]{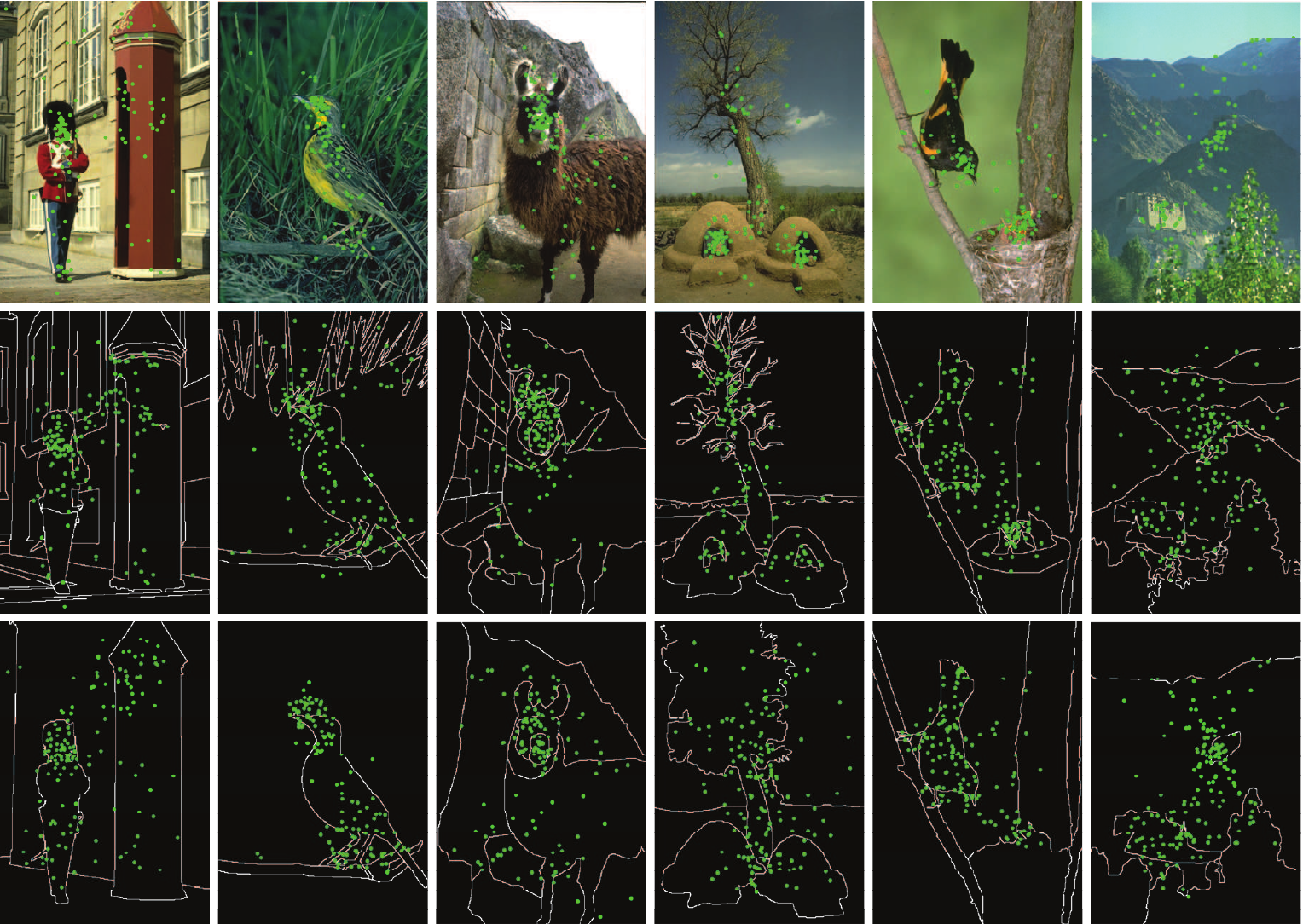}
\end{center}
\caption{Examples of fixation points (green dots) on natural scenes and the corresponding line drawings. From top to bottom row: the fixations from 10 subjects when viewing the natural scenes, the line drawings $C_{max}$ and $C_{min}$, respectively.}
\label{FigFix}
\end{figure*}

In addition, we further consider the radii of regions to boost the estimate of closure degree. Let $ R_i(x,y), i \in [1,D] $ denotes the radii of all the considered directions referring to point $ (x,y) $. Subsequently, we compute the final closure degree of $ (x,y) $ as

\begin{equation}
{S_{cc}}\left( {x,y} \right) = \frac{{C\left( {x,y} \right)}}{{\tilde R(x,y) + S(x,y)}}
\label{eqn2}
\end{equation}
where, $ \tilde R(x,y) $ is the mean radius and $ S(x,y) $ is the standard deviation of all the radii along $ D $ directions when considering point $ (x,y) $. With the definition in Equation (\ref{eqn2}), the small regions (with low $ \tilde R(x,y) $) have the higher closure degrees than the larger regions. Meanwhile, low $ S(x,y) $ means the region towards a circle which attracts more attention. 

In order to evaluate the contribution of contextual guidance in fixation prediction, we build a simple model to improve the performance of fixation prediction when integrating the spatial layout information into the classical bottom-up methods. In detail, we employ a Gaussian mixture model (GMM)  \cite{permuter2003gaussian} to fit the closure degree map obtained with Equation (\ref{eqn2}) and find the $K$ main components (we set $ K=5 $ in the experiments). The line drawing guided prior is obtained by combining multiple GMM components ($ {G_i}(x,y) $) with weights ($ {\omega _i} $) 

\begin{equation}
{W_{GMM}}(x,y) = \sum\limits_{i = 1}^K {{\omega _i} \cdot {G_i}(x,y)}
\label{eqn3}
\end{equation}

\begin{equation}
{\omega _i} = \omega _{porp}^i \cdot \omega _{ellip}^i \cdot \omega _{dis}^i
\label{eqn4}
\end{equation}
where, each component is weighted by the component proportion ($ \omega _{porp}^i $) , ellipticity ($ \omega _{ellip}^i $), and the distance weight ($ \omega _{dis}^i $). In detail, $ \omega _{porp}^i $ is the component proportion from GMM fitting. $ \omega _{ellip}^i $ is related to the ellipticity which is computed with the long and short radius the $ i^{th} $ Gaussian component along its two main axis. It means that the rounder component has a greater weight. $ \omega _{dis}^i $ is related to the distance of the center of the $ i^{th} $ Gaussian component to the image center, which indicates that the component nearer to image center has a greater weight.

Finally, the combined saliency map ($ S_{com} $) can be obtained as
\begin{equation}
{S_{com}}\left( {x,y} \right) = {S_{bottom-up}}(x,y){\kern 1pt} {\kern 1pt} {\kern 1pt} {\kern 1pt} {\kern 1pt}  \cdot {\kern 1pt} {\kern 1pt} {\kern 1pt} {\kern 1pt} {\kern 1pt} {W_{GMM}}(x,y)
\label{eqn5}
\end{equation}
where, $ S_{bottom-up}(x,y) $ is a saliency map obtained by a low-level saliency method, like IT \cite{itti1998model}. With Equation (\ref{eqn5}), we simply obtain the line drawing (closure cue) guided visual saliency map. 

\section{Results}
\subsection{Human fixation analysis}
Compared to the natural scenes, the line drawings of an image excludes the basic low-level features such as the color and luminance. However, what is the difference between the distribution of fixations on the line drawings and that on the natural image of a same scene?  Figure \ref{FigFix} shows several examples of the fixation distribution on the natural images and the corresponding line drawings. In this figure we display all the fixation points from the 10 subjects on each visual stimulus. Note that we removed the first fixation point of each subject in order to eliminate the center bias considering that the visual cue of cross marker was presented at the center of the gray screen.

From Figure \ref{FigFix}, we can clearly see that the fixations on different stimuli (natural scenes and line drawings) of the same scenes are quite similar. Most of the fixation points (green dots) locate on the dominant objects in both the natural scenes and the line drawings. Especially, without the color and surface information, visual attention is also attracted by the objects in the line drawings, which are defined by the shapes and the Gestalt features of local regions. Note that without the specific surface features inside the objects, the fixations on the line drawings are distributed more dispersively than that on the natural scenes. This suggests that the visual attention is attracted by the structure of scenes and the shape features of the potential objects that are outlined by the dominant contours in the line drawings. When we are viewing the natural scenes, the local feature contrasts can further gather our attention to some salient regions of the visual objects. This also indicates that the layout and the structure of scenes may guide our attention to the dominant objects for further visual processing.

\begin{figure}[htbp]
\begin{center}
\includegraphics[width=9cm]{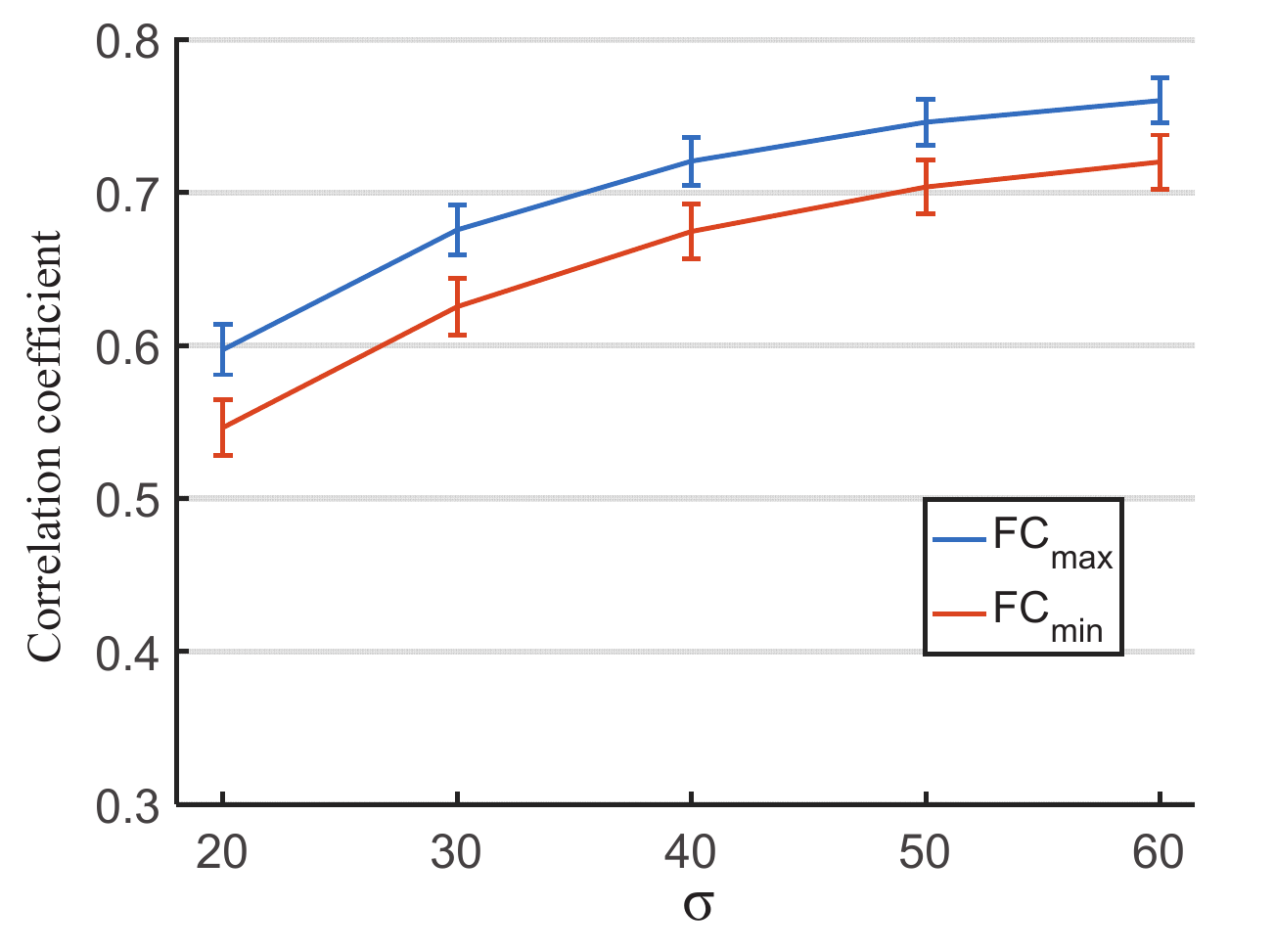}
\end{center}
\caption{Correlation coefficients between the fixation distributions on the line drawings ($C_{max}$ and $C_{min}$) and the natural images with varying scales ($ \sigma $ in pixel) of Gaussian blurring. The fixation distributions on the line drawings of $C_{max}$ and $C_{min}$ are denoted by $FC_{max}$ and $FC_{min}$ respectively. The error bars are the 95\% confidence intervals.}
\label{FigFixCC}
\end{figure}

For qualitative comparison, we also calculated the correlation coefficient between the distribution of fixations on the line drawings and that on the natural images. Considering that the fixation points are extremely sparse on the stimuli, various scales of Gaussian blurring were used to obtain the spatial distribution of fixations. The correlation coefficients (CC) were computed between the blurred distributions of fixations on the line drawings and the natural scenes, i.e., 
\begin{equation}
CC(f_{ld},f_{ns}) = \frac{{cov(f_{ld},f_{ns})}}{{\sigma_{ld}\sigma_{ns}}}
\label{eqnA1}
\end{equation}
where, $cov()$ is the covariance, $\sigma_{ld}$ and $\sigma_{ns}$ are the standard deviation of blurred distributions of fixations on the line drawings ($f_{ld}$) and the natural scenes ($f_{ns}$). Figure \ref{FigFixCC} shows the correlation coefficients with varying blurring scales. The fixation distributions on the line drawings of $C_{max}$ and $C_{min}$ are denoted by $FC_{max}$ and $FC_{min}$ respectively. We can see that compared to the line drawings $C_{min}$ containing less details, the line drawings $C_{max}$ containing more details provide more visual information, and the fixation distribution on $C_{max}$ is slightly closer to that on the natural images. 

\begin{figure*}[h!]
\begin{center}
\includegraphics[width=10cm]{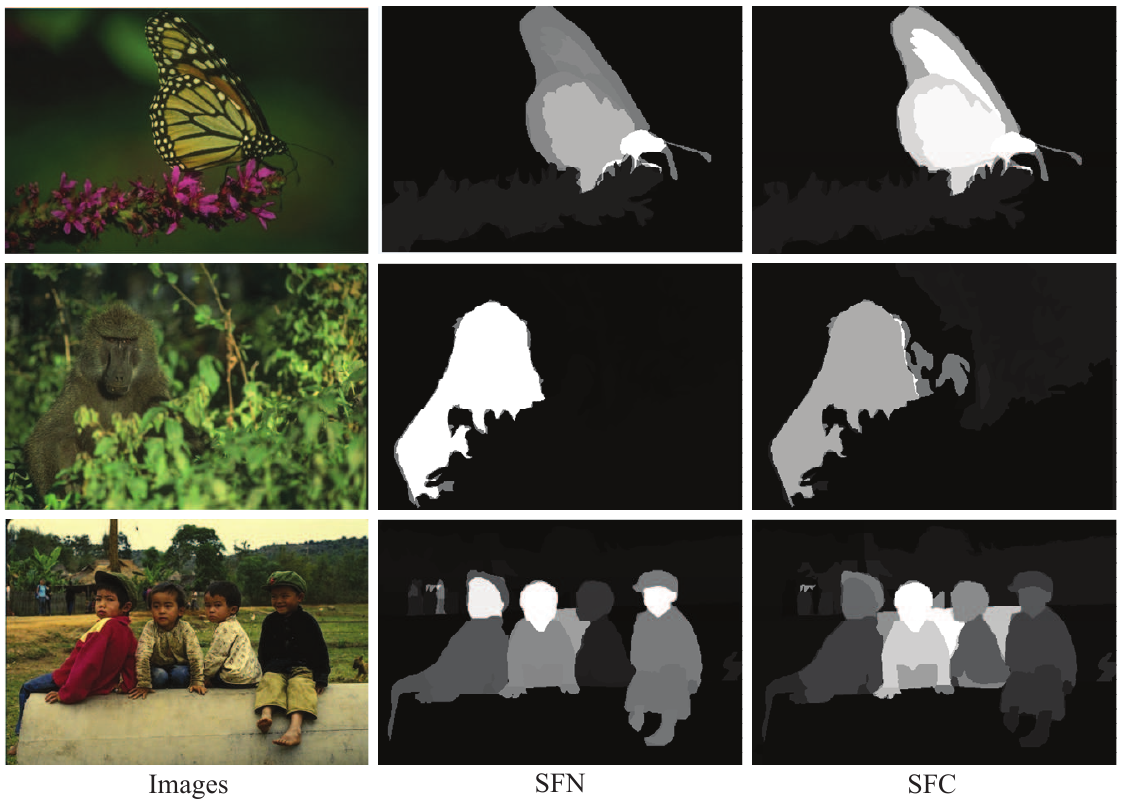}
\end{center}
\caption{Allocating saliency score to each region with fixation density. $SFN$ denotes the hierarchical salient objects generated based on the human-marked segmentations and fixations on the natural images, while $SFC$ denotes the hierarchical salient objects generated based on the human-marked segmentations and fixations on the contours (i.e., line drawings). Brighter regions have higher saliency scores.}
\label{FigObj}
\end{figure*}

Figure \ref{FigFixCC} indicates the high correlation (the correlation coefficients are around 0.7) of fixation distribution between the natural scenes and their line drawings. However, it is easily understood that the fixation locations of different subjects are usually not exactly consistent with each other even they are viewing the same objects. Therefore, we suppose that different subjects are attending to the same objects if their fixation points locate within the different parts of the same objects. In order to evaluate the object-based consistence between the visual attention on the natural scenes and the line drawings, we first generated the salient regions based on the human-marked segmentation provided by BSDS500 as follows. For each segment, we computed the density of the human fixation points as its saliency level. Thus, if a segment (or a region) attracts more fixation points, it will obtain a higher saliency score. For each scene, the two salient object maps generated based on the two line drawings (i.e., $C_{max}$ and $C_{min}$) and the corresponding fixations were linearly summarized together and integrated into the final map containing the hierarchical salient objects. Similarly, we obtained the hierarchical salient objects based on the human-marked segmentations and fixations collected on the original images. Figure \ref{FigObj} shows several examples of hierarchical salient objects generated from the fixations on the natural images and the line drawings.

\begin{figure*}[h!]
\begin{center}
\includegraphics[width=10cm]{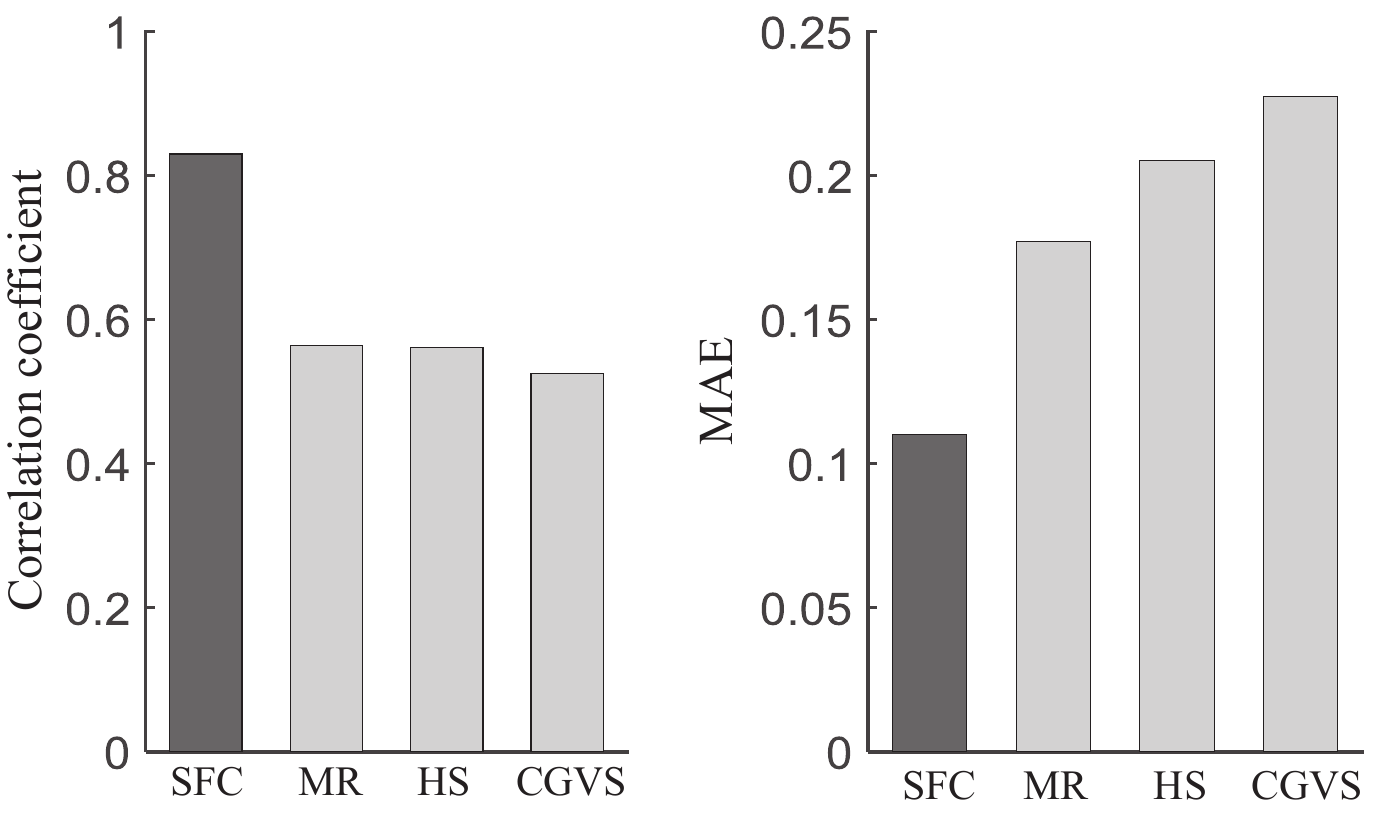}
\end{center}
\caption{Correlation coefficients and MAEs of the salient regions based on the hierarchical salient objects. The baseline is the salient objects generated with the fixations on the natural scenes ($SFN$), while $SFC$ denotes the hierarchical salient objects generated based on the human-marked segmentations and fixations on line drawings}
\label{FigObjCC}
\end{figure*}

In this work, the object-based consistence of visual attention was evaluated based on the hierarchical salient objects using two metrics: correlation coefficient \cite{borji2013analysis} and Mean Absolute Error (MAE) \cite{Perazzi2012Saliency, yang2016unified}. As indicated in Figure \ref{FigObjCC}, with the salient objects generated with the fixations on the natural scenes ($SFN$) as baseline, the salient objects generated with the fixations on the line drawings ($ SFC $) obtain clearly higher correlation coefficient and lower MAE than some representative computational models (including MR \cite{yang2013saliency}, HS \cite{yan2013hierarchical}, and CGVS \cite{yang2016unified}). Actually, this is an obvious conclusion considering that these hierarchical salient objects were marked based on the fixations from human subjects. However, this experiment indeed suggests the high consistence of visual attention when we are viewing the natural images and the line drawings of the same scenes. Although without the color and surface information, human subjects can also execute efficient visual searching by only employing the limited scene structure information.

\subsection{Guidance cues}
With the absence of color and luminance, how does the line drawings guide our visual attention? Some studies have revealed the role of scene structure for visual guidance in specific visual scenes or tasks. For example, previous studies have shown that the vanishing point of a road plays an important role in traffic scenes especially with the driving task \cite{deng2016does}. However, most of the stimuli used in this experiment are task-irrelevant general images, each of which usually contains at least one dominant object. Therefore, we believe that the shape features of regions and scene layout mainly contribute to predict potential objects.

We tested ten shape-related common features and evaluated their correlation with the fixation distributions. These shape features were extracted from each segment of each image, which are listed in Table \ref{T1}.  These features can reflect different characters of regions. For example, the regions with higher degree of \textit{Closure} are usually related to meaningful objects of higher probability.  \textit{Perimeter Ratio} carries the continuity of the boundary of a region. Higher \textit{Perimeter Ratio} indicates higher irregularity of a region.  It has been verified that some of these features are helpful for the salient object detection \cite{li2014secrets}. However, in this work, we further analysed the contribution of individual features to the specific task of visual attention.

\begin{table*} \small
  \centering
  \caption{Ten shape-related common features used in this study and their correlation with the fixation distributions.}
  \label{T1}
  \begin{tabular}{ p{2.5cm} p{11cm} p{1.0cm} p{1.0cm} }  
  \hline
  Features & Description & $C_{max}$ & $C_{min}$ \\
  \hline
  Area Ratio  & The ratio of number of pixels in the region to total pixels in the image  & -0.459 & -0.483  \\
  Centralization  & The mean distance of pixels in the region to the center of image, normalized by the image size  & -0.409 & -0.301  \\
  Perimeter Ratio  & The ratio of perimeter of the region to perimeter of the image  & -0.510 & -0.526  \\
  MinAxis/MajorAxis & The length of the major axis divided by the length of the minor axis of the region  & -0.072 & 0.072  \\
  Eccentricity  & The eccentricity of the ellipse that has the same second-moments as the region  & -0.067 & -0.065  \\
  Orientation & The orientation of the major axis of the ellipse that has the same second-moments as the region & 0.013 & -0.006  \\
  EquivDiameter  & The ratio of diameter of a circle with the same area as the region to the diagonal length of image  & -0.609 & -0.601  \\
  Solidity  & The proportion of the pixels in the convex hull that are also in the region  & -0.047 & -0.013  \\
  Extent  & The ratio of pixels in the region to pixels in the total bounding box & -0.066 & -0.234 \\
  Closure  & The closure score of a region is defined in Equation(\ref{eq02})  & 0.632 & 0.736  \\
  \hline
  \end{tabular}
\end{table*}

\begin{figure*}[h!]
\begin{center}
\includegraphics[width=16cm]{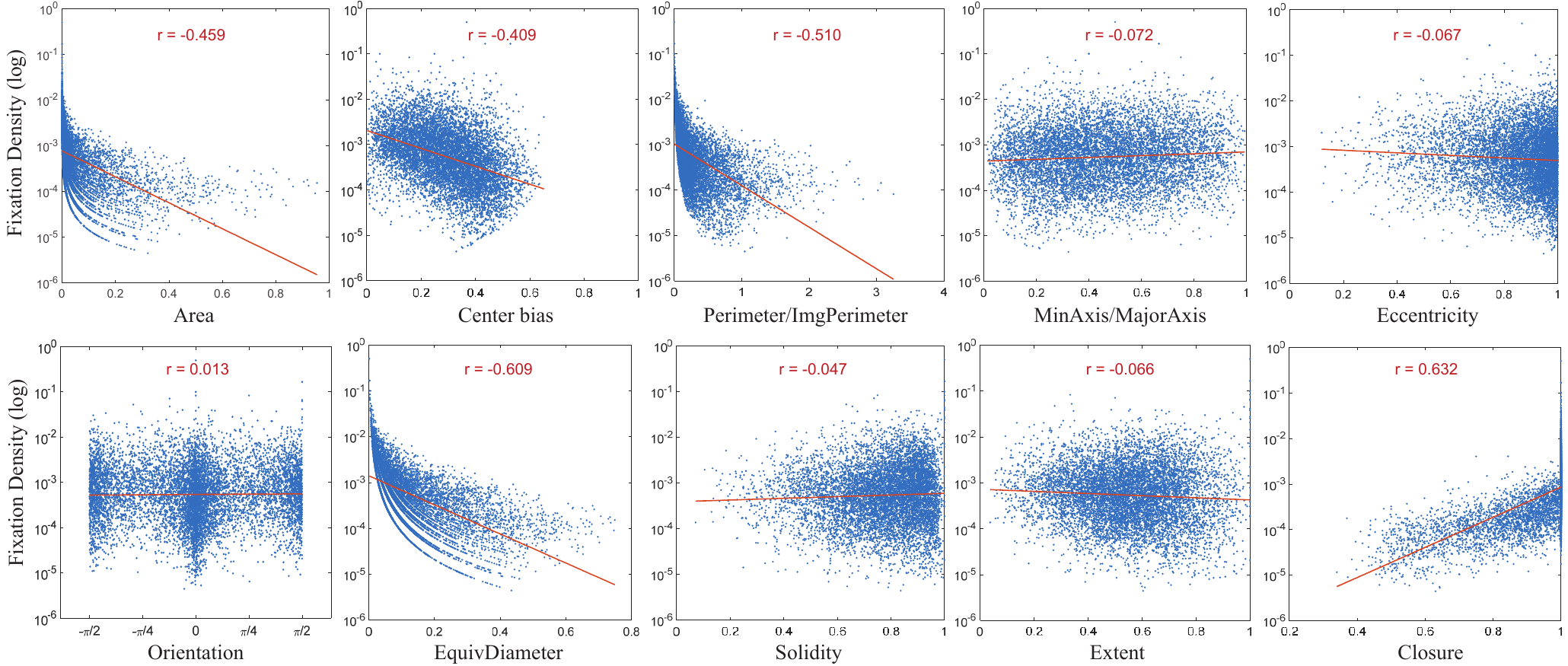}
\end{center}
\caption{Correlations between various shape features and the saliency levels of each region based on $C_{max}$. The number listed for $r$ in the top of each panel is the linear correlation coefficient between the feature and the saliency values (in log scale).}
\label{FigShapes}
\end{figure*}

Figure \ref{FigShapes} lists the relations between the saliency values (which are defined by the fixation density) and the corresponding shape cues of each segment of $C_{max}$. In order to clearly show the distributions, the saliency values are shown in log scale in Figure \ref{FigShapes}. The relations between the saliency values and the corresponding shape cues of each segment on $C_{min}$ are similar with that on $C_{max}$. Table \ref{T1} also lists the correlation coefficient of saliency values with each shape cue on $C_{max}$ and $C_{min}$. We can find that the feature of \textit{Closure} contributes most to visual attention among all the cues considered here. In addition, the contributions of \textit{Closure}, \textit{EquivDiameter} and \textit{Perimeter Ratio} are higher than that of the well-known center bias (\textit{Centralization}). In contrast, some features like \textit{Orientation} contribute little for object detection. It make senses that most of the regions in the natural scenes tested in this work are distributed on the 0 degree and 90 degree orientations (the bottom-left panel in Figure \ref{FigShapes}), but \textit{Orientation} cannot provide enough information to distinguish between object and non-object regions.

It has been proved that the closure, as the most important one among the considered features, plays a pivotal role for describing an object proposal \cite{yu2016computational}. Here we further evaluated the contribution of closure feature in the specific task of visual searching. Firstly, we define a score to measure the degree of closure of a region based on the human-marked segmentation. Let $NB^i$ denote the number of intersection between the boundary pixel of segment $i$ and the border pixel set of the image and $NR^i$ denote the number of all boundary pixels of segment $ i $. Then, the closure score of contour pixels on segment $ i $ is defined as
\begin{equation}
C{S_i} = 1 - \frac{{N{B^i}}}{{N{R^i}}}
\label{eq02}
\end{equation}

We denote the set of fixations as $ S_f $ and the set of contour pixels as $ S_c $. In addition, we define the closed region as the segment with closure score higher than 0.9  (i.e, $ CS_i > 0.9 $), and the set of pixels on closure contours and in the closure regions as $S_{cc}$. 

Then, we define four metrics as follows: (1) the percentage of the fixations around the contours in all fixations ($PoF$); (2) the percentage of the contours around the fixations in all contour pixels ($PoC$); (3) the percentage of the fixations around and in the closed regions in the fixations around all contours ($PoFC$); (4) the percentage of the closed regions in the whole image ($PoCC$). These four metrics can be computed as
\begin{equation}
PoF = \frac{{{D_N}({S_c}) \cap {S_f}}}{{{S_f}}}
\label{eq03}
\end{equation}

\begin{equation}
PoC = \frac{{{S_c} \cap {D_N}({S_f})}}{{{S_c}}}
\label{eq04}
\end{equation}

\begin{equation}
PoFC = \frac{{{D_N}({S_{cc}}) \cap {S_f}}}{{{S_f}}}
\label{eq05}
\end{equation}

\begin{equation}
PoCC = \frac{{{D_N}({S_{cc}})}}{{W \times H}}
\label{eq06}
\end{equation}
where $ D_N(S) $ denotes the dilatation operator with a size of $ N \times N $ pixels applied on $ S $ (a binary map), $ W $ and $ H $ represent respectively the width and height (in pixels) of the image. 

\begin{figure*}[h!]
\begin{center}
\includegraphics[width=13cm]{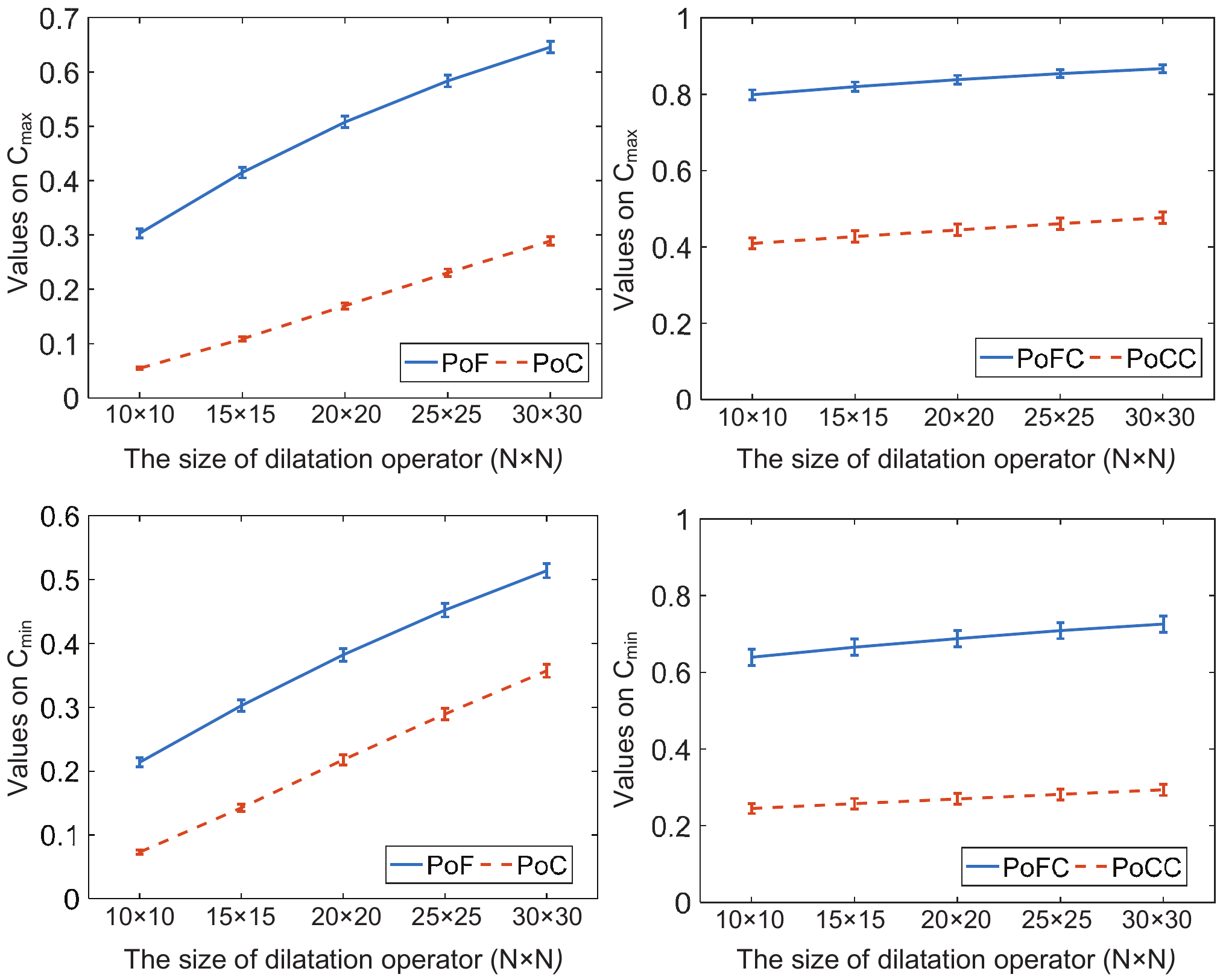}
\end{center}
\caption{The multiple metrics under dilatation operation with varying operator size of $ N \times N $ pixels on the line drawings of $C_{max}$ (top row) and $C_{min}$ (bottom row). Left: $PoF$ and $PoC$; Right: PoFC and PoCC. The error bars are the 95\% confidence intervals.}
\label{FixCon}
\end{figure*}

Figure \ref{FixCon} illustrates graphically the metrics with varying $ N $ values on the line drawings of $C_{max}$ and $C_{min}$. From Figure \ref{FixCon} (left), $PoF$ only achieves around 0.5 even with a large $N$, which indicates that less than half of the fixation points locate near the contour lines and more than half of the fixation points locate in the non-contour regions (without any local contrast). This suggests that there are some higher-order visual cues beyond local contrast (contours) for guiding visual attention. In addition, although contour lines attract around half of the visual fixations, $PoC$ further shows that these fixations locate around only 30\% of all contour pixels. This means that part of contour lines plays a more important role than others in guiding visual attention. 

A simple suppose is that visual objects are commonly defined by the closed regions in line drawings and visual attention usually focuses on the potential visual objects. Examples shown in Figure \ref{FigFixCC} and the results listed in Table \ref{T1} suggest that closed regions attract more visual fixation. In order to further verify this suppose, we analyzed the metrics of $PoFC$ and $PoCC$ with varying $N$ values on the line drawings of $C_{max}$ and $C_{min}$. Figure \ref{FixCon} (right) indicates that around 80\% of the fixations locate within or around the closed regions (high $PoFC$), although the closed regions cover only a small percentage of area in the whole scene (low $PoCC$).

\subsection{Model evaluation}
\label{S33}
\begin{figure}[htbp]
\begin{center}
\includegraphics[width=8cm]{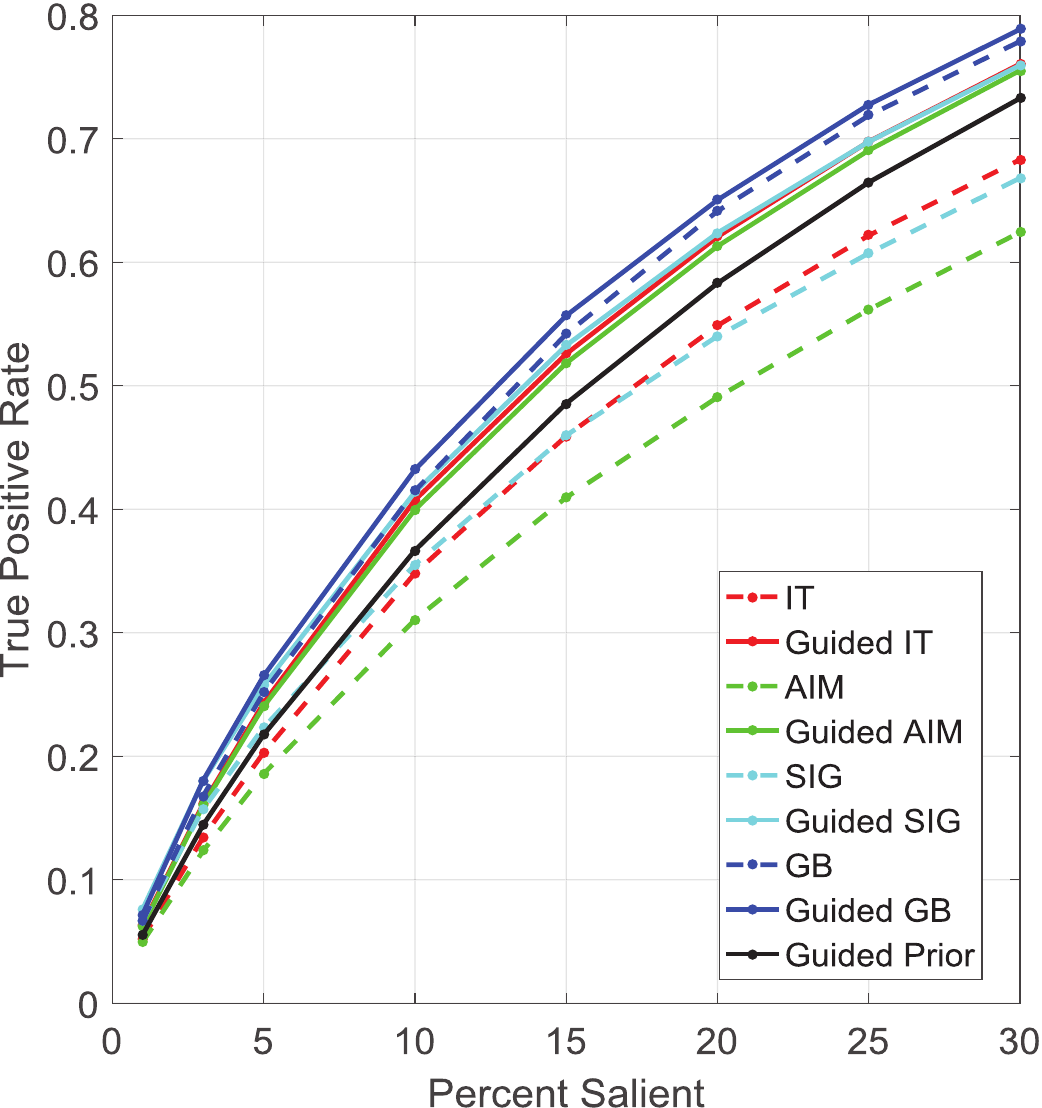}
\end{center}
\caption{Performance comparison between the bottom-up models and the improved models when integrating the guided information from line drawings.}
\label{FigROCs}
\end{figure}

\begin{figure*}[htbp]
\begin{center}
\includegraphics[width=13cm]{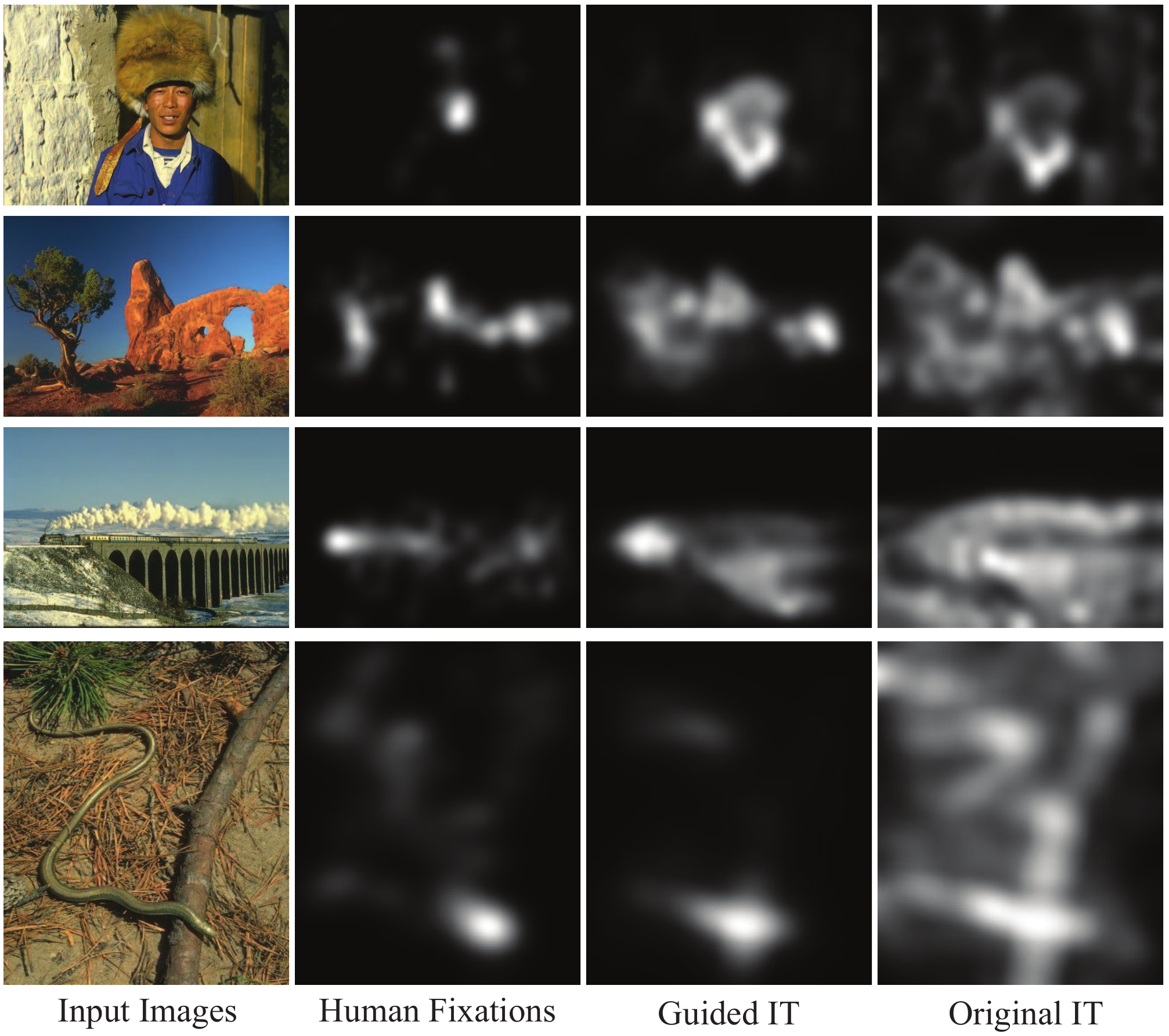}
\end{center}
\caption{Examples of saliency maps predicted by the original IT and the improved model (Guided IT).}
\label{FigSMs}
\end{figure*}

In this experiment, we employed several representative classical models including FIT-based model (IT) \cite{itti1998model}, AIM \cite{bruce2006saliency}, SIG \cite{hou2012image}, graph-based model (GB) \cite{harel2006graph}. To evaluate the performance of multiple models for the task of fixation prediction, we employed a metric of receiver operating characteristic (ROC) curve widely used in the field of computer vision \cite{borji2013analysis, bruce2006saliency}. In this experiment, we used the revised version of ROC \cite{judd2009learning}, which focuses mainly on the true positive rate, i.e., the proportion of actual fixations that are correctly identified as such, and the human fixation on the natural scenes is regarded as the ground truth. As shown in Figure \ref{FigROCs}, the fixation prediction performances of all the considered bottom-up models are significantly improved after integrating the guided information from line drawings using Equation (\ref{eqn5}). For example, the IT model is a classical bottom-up model, which predicts the fixations from natural images by combining the contrast features of color, luminance and orientation \cite{itti1998model}. Therefore, with the absence of scene structure feature, IT model mainly detects some regions with high local contrasts while missing the meaningful object regions (e.g., the inner surface of objects). However, line drawings can provide additional scene structure and regional shape information, which serve to guide the visual attention to focus on the inner of potential objects. Therefore, the closure prior from the line drawings that represent the important region information can remarkably improve the performance of IT. Figure \ref{FigSMs} shows several examples of saliency maps predicted by the original IT and the improved model (Guided IT).

\section{Discussion}
Classical theories of visual processing view the brain as a stimulus-driven device, in which visual information is extracted hierarchically along the visual pathway \cite{treisman1980feature}. However, numerous recent neurophysiological and psychological studies support that a coarse-to-fine mechanism plays an important role in various visual tasks \cite{mermillod2005coarse}, such as stereoscopic depth perception \cite{menz2003stereoscopic}, temporal changes \cite{hegde2008time}, object recognition in context \cite{bar2004visual}, etc. In addition, Oliva et al. reveal that the scale usage of visual information is flexible and can provide the information required depending on the task constraints \cite{oliva1997coarse}. On the other hand, it has been proven that the coarse-to-fine strategy can contribute to various computer vision applications, e.g., contour detection \cite{zeng2011center} and image segmentation \cite{yao2015real}. These researches further confirm the efficiency of the coarse-to-fine mechanism in visual information processing from the viewpoints of computational modeling. 

It is widely accepted that visual processing is an active and highly selective process \cite{eckstein2011visual}, which reflects the dynamic interaction between the bottom-up feature extraction and the top-down constraints \cite{engel2001dynamic, wolfe2011visual}. This means visual information processing is not purely stimulus-evoked, but constrained by top–down influences that strongly shape the intrinsic dynamics of cortical circuits and constantly create predictions about the forthcoming information \cite{engel2001dynamic, gilbert2013top, hopf2000neural}. For example, with object recognition in context, context-based prediction makes the task of object recognition more efficient \cite{bar2004visual}. These studies support that context information, acting as coarse and global scene information, can be rapidly extracted and used to facilitate the object recognition in complex scenes \cite{bar2004visual, bar2006top}.  

As an important aspect of scene perception, visual attention is also considered as a process that encompasses many aspects central to human visual and cognitive function \cite{eckstein2011visual}. Knudsen also proposed a famous conceptual framework that indicates the attention combined contribution of four distinct processes: working memory, competitive selection, top-down sensitivity control, and automatic filtering for salient stimuli \cite{knudsen2007fundamental}. Computational modeling of visual searching has also demonstrated the contribution of information from different levels of perception \cite{torralba2006contextual}. 

According to the scales of features, visual information can be coarsely divided into three levels: low, middle, and high levels. Firstly, at the low-level scale, classical bottom-up frameworks (e.g., Koch, Itti et al. \cite{koch1987shifts, treisman1980feature, itti1998model}) have shown that local contrast in various feature channels can attract visual attention. These models are usually based on the local filtering and obtain stimuli-driven saliency map without taking into consideration the behavioral goals of searching \cite{itti2001computational}. Secondly, at the region scale (mid-level), perceptual organization principles which describe how basic visual features are organized into more coherent units  \cite{wagemans2018perceptual} (e.g., Gestalt \cite{Koffka1935Principles})  could guide the visual searching. For example, the regions that match some specific principles (e.g., closure, continuity, symmetry, etc.) usually indicate special visual objects that are more interesting for human visual tasks \cite{kootstra2008paying,dickinson2018visual}. Finally, at the scene scale (high-level), scene layout and structure will be important guidance to predict where the interesting objects present in the current scene \cite{wolfe2011visual, eckstein2017probabilistic}. With the guided search theory, the high-level scene semantic and gist could provide important spatial cues for targets, which will facilitate the binding of low-level features and speed-up the visual searching \cite{wolfe2011visual, oliva2001modeling, torralba2006contextual, deng2016does}. These principles are usually solidified in our memory as certain general knowledge obtained with learning from our daily life.

In this paper, we focus on the role of line drawings in guiding visual attention. Generally, line drawings of natural scenes provide two structure related information, i.e., the shape feature of region (mid-level) and the layout of scene (high-level). Therefore, line drawings can represent two types of contextual sources. On the one hand, line drawings segment an image into various perceptional regions (before the perception of specific objects) that may be used by the visual system to rapidly construct a rough sketch of the scene structure in space \cite{oliva2006building}, and also contribute to fill-in the surface of regions \cite{zweig2015representation}. At the level of region, visual object shapes usually obey some general principles that can be considered as certain general prior or knowledge. For example, Gestalt principles describe some general principles of visual perception \cite{Koffka1935Principles}. As for the visual attention, the shapes of regions matching some Gestalt principles (e.g., closure, continuity, etc.) could attract more visual attention because they represent meaningful visual objects with higher probability. Moreover, geometrical properties of regions have been strongly proven to contribute to early global topological perception \cite{chen2005topological}. 

On the other hand, line drawings provide the coarse scene structure. The visual guided search theory proposed by Wolfe et al. \cite{wolfe2011visual} suggests that scene global information (including scene spatial layout) can be transferred rapidly to high visual cortexes via the so-called non-selective visual pathway. Line drawings could give the rough layout of surfaces in the space and provide the basis for scene-based guidance of visual searching. Physiological evidence shows that the global contours delineating the outlines of visual objects may be responded quite early (perhaps via a special pathway) by the neurons of high cortexes, which, although producing only a crude signal about the position and shape of the objects, can provide sufficient feedback modulation to enhance the contour-related responses at lower levels and suppress the irrelevant background inputs \cite{Chen2014Incremental}. In addition, coarse contour is low frequency visual information \cite{zeng2011center}, which carries coarse scene information to provide enough signals to infer the scene structure and layout \cite{bar2004visual, bar2006top}. 

To summarize, visual search should be a unified framework that integrates information from different sources in the brain \cite{schutt2019disentangling}. As important guidance cues, line drawings of natural scenes provide a rapid representation of the scene structure and regional shape. Coarse scene layout may be not enough for the object identifying tasks, but it can provide efficient cues to predict where the potential target is and contribute to the active vision system to achieve high efficient visual searching and scene perception \cite{eckstein2011visual}.

\section{Conclusion}
In this paper, we collected a dataset of human fixations on the natural scenes and the corresponding line drawings, which are useful for analyzing the mechanisms of visual attention at the object or shape level. Our experiments reveal a high correlation between the distribution of fixations on the natural images and that on the line drawings of the same scenes. In addition, we systemically analyzed the effects of various shape-based visual cues in guiding visual attention. The results suggest that the closed regions have higher possibility to form meaningful objects and may attract most of our visual fixations. Finally, the computational model further verifies that the information of line drawings can be used to significantly improve the fixation prediction performance of classical bottom-up methods on natural scenes.

In conclusion, we suggest that the cortexes involved in visual attention or visual search should be decomposed into not only various parallel feature channels (such as the IT model \cite{itti1998model}), but also various hierarchical levels including low-, mid-, and high-levels. At the same time, the information from various levels plays different roles in visual searching. Therefore, our future work will be extended as following aspects: (1) to build a computational model to predict the scene layout, which will be a powerful complement to the low-level visual features for visual object searching; (2) to further study how the scene structure information guides our visual attention; (3) to integrate the guidance information from various scales and build a unified framework for visual guided searching tasks combining the low-, mid-, and high-level guided cues.

\section*{Acknowledgment}
This work was supported by Natural Science Foundations of China under Grant 61703075, Sichuan Province Science and Technology Support Project under Grant 2017SZDZX0019, and Guangdong Province Key Project under Grant 2018B030338001.


\bibliographystyle{IEEEtran}
\bibliography{Refs}



\end{document}